\pdfoutput=1

\documentclass[11pt]{article}

\usepackage{acl}

\usepackage{times}
\usepackage{latexsym}

\usepackage[T1]{fontenc}

\usepackage[utf8]{inputenc}

\usepackage{microtype}

\usepackage{inconsolata}

\usepackage{graphicx}
\usepackage{amsmath}
\usepackage{amsfonts}
\usepackage{booktabs}
\usepackage{multirow}
\usepackage{xcolor}
\usepackage{subcaption}

\usepackage{booktabs}
\usepackage{array}
\usepackage[table]{xcolor}
\usepackage{makecell}
\usepackage{enumitem}

\definecolor{demoRow}{HTML}{F6FAF3}
\definecolor{oursRow}{HTML}{EAF3FF}
\definecolor{posGreen}{HTML}{007A2F}
\definecolor{negRed}{HTML}{C43C35}
\definecolor{weakRow}{HTML}{F7F7F7}
\definecolor{strongDemoRow}{HTML}{DFF1E6}
\definecolor{softGray}{HTML}{6F6F6F}

\newcommand{\gain}[1]{\textcolor{posGreen}{+#1}}
\newcommand{\drop}[1]{\textcolor{negRed}{#1}}
\newcommand{\gainx}[2]{\textcolor{posGreen}{+#1 {\scriptsize (#2$\times$)}}}
\newcommand{\na}{--}
\newcommand{\group}[1]{\multicolumn{14}{l}{\textbf{#1}}}

\usepackage{tikz}
\usetikzlibrary{arrows.meta, positioning, calc}

\definecolor{brickred}{RGB}{178,34,34}

\usepackage{amssymb}
\usepackage{tcolorbox} 
\tcbset{
    colback=gray!10,   
    colframe=black,    
    boxrule=0.5pt,     
    arc=2mm,           
    left=2mm, right=2mm, top=1mm, bottom=1mm 
}
\title{When Does Demographic Information Help?\\ Data and Modeling Regimes for Perspective-Aware Hate Speech Detection}

\author{Weibin Cai \\
  Data Lab, EECS Department \\
  Syracuse University \\
  \texttt{weibin44@data.syr.edu} \\\And
  Reza Zafarani \\
  Data Lab, EECS Department \\
  Syracuse University \\
  \texttt{reza@data.syr.edu} \\}

\begin{document}
\maketitle
\begin{abstract}
Demographic information is often used to model annotator perspectives in subjective tasks such as hate speech detection, but its benefit is inconsistent: \textit{it  improves performance in some settings and behaves as noise in others}. 
This paper asks \textbf{\textit{when demographic features help}}. 
We analyze demographic gain as a function of both \textbf{data split} properties and \textbf{modeling frameworks}. 
For data splits, we measure annotator disagreement, namely how often annotators assign different labels to the same example, along with training size and train--test demographic coverage.
We find that demographic gains concentrate in regimes with  \textbf{\textit{low training disagreement}}, \textbf{\textit{high test disagreement}}, \textbf{\textit{fine-grained ambiguity measurement}}, \textbf{\textit{sufficient training data}}, and \textbf{\textit{greater demographic overlap}}. 
Motivated by these regimes, we introduce a gated demographic residual model that treats demographics as a selective adjustment to text-only predictions. 
Experiments on MHS and POPQUORN show that this design is effective, especially on high disagreement or low confidence examples. 
Overall, our results suggest that demographics should not be assumed useful by default; their value depends jointly on the data regime and the modeling framework.

\end{abstract}

\begin{figure}[t]
    \centering
    \includegraphics[width=\linewidth]{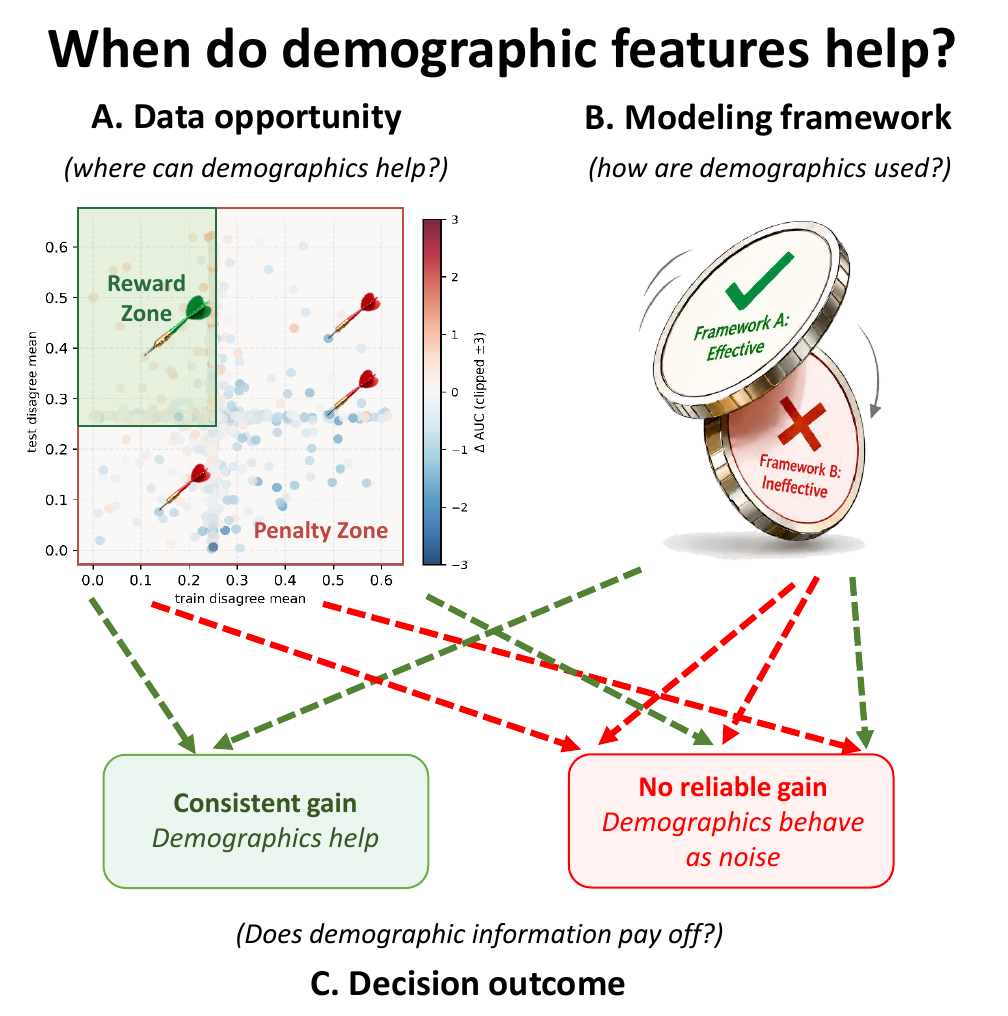} 
    \caption{Data split and modeling frameworks jointly determine whether demographic information helps. Demographic gains are most likely when the split falls in a favorable region and the framework incorporates demographics in an effective way. Otherwise, demographics may provide no reliable gain or behave as noise.}
    \vspace{-5mm}
    \label{fig:demographic-features-help}
\end{figure}


\begin{table*}[t]
\centering
\small
\setlength{\tabcolsep}{6pt}
\begin{tabular}{lccccccl}
\toprule
Setting
& \makecell{Reward-zone\\split?}
& \makecell{Effective\\framework?}
& \makecell{Text\\AUC}
& \makecell{+Shuffled demo.\\AUC}
& \makecell{+Real demo.\\AUC}
& $\Delta$AUC
& \makecell{Are demographics useful?} \\
\midrule

B--B
& No & No
& \textbf{90.41} & 90.09 & 90.40 & \drop{-0.01}
& \textbf{\textcolor{negRed}{No}} \\

B--A
& No & Yes
& \textbf{92.26} & 91.91 & 92.25 & \drop{-0.01}
& \textbf{\textcolor{negRed}{No}} \\

A--B
& Yes & No
& 78.03 & 77.93 & \textbf{78.21} & \gain{0.18}
& \textbf{\textcolor{softGray}{Weak / inconclusive}} \\

\rowcolor{strongDemoRow}
A--A
& \textbf{Yes} & \textbf{Yes}
& 64.73 & 64.54 & \textbf{66.01} & \textbf{\gain{1.28}}
& \textbf{\textcolor{posGreen}{Yes, clear gain}} \\

\bottomrule
\end{tabular}
\caption{
Demographic features become useful when both conditions align.
The setting column denotes the paired split--framework configuration, e.g., A--B uses Split A with Framework B.
A reward-zone split has low training disagreement and high test disagreement, while an effective framework directly models demographic information.
$\Delta$AUC is computed as the AUC difference between real demographics and the text-only baseline under the same setting.
Shuffled demographics serve as a negative control. 
Full results are reported in Appendix~\ref{app:intro_case}.
}
\vspace{-4mm}
\label{tab:demo_split_framework_summary}
\end{table*}

\section{Introduction}

Determining whether a text is hateful is often subjective and perspectival~\cite{frenda2025perspectivist}. 
Traditional hate speech annotation typically collapses multiple annotations into a single majority-vote label, which can introduce bias and obscure meaningful disagreement among annotators~\cite{sap2019social}. 
Data perspectivism instead treats annotator disagreement as informative signal rather than noise to be removed~\cite{cabitza2023toward}. 
Following this view, prior work has examined how annotator demographics and identities shape hate speech judgments, including factors such as race, gender, and political leaning~\cite{sap2022annotators,lee2023exploring,davani2024d3code}. 
Another line of work models annotator or group information to capture variation in subjective judgments, often assuming that demographic information provides a useful modeling signal~\cite{xu2025modeling,wan2023everyone,fleisig2023majority}.

However, demographics are not always helpful:
they can be sparse, noisy, or harmful when injected blindly.
Some studies find that adding demographic features yields only marginal gains or even hurts performance~\cite{sorensen2025value,orlikowski2025beyond,tahaei2025demographic,yin2023annobert}; one explanation is the ecological fallacy, where group-level patterns do not necessarily explain individual behavior~\cite{orlikowski2023ecological}. 
Other work reports clear benefits from modeling demographic or annotator variation~\cite{fleisig2023majority,jaggi2024accurate,gordon2022jury,xu2025modeling}. 
This mixed evidence raises a basic question: \textit{when do demographics help?}

We argue that the answer depends on two sources of variation, as illustrated in Figure~\ref{fig:demographic-features-help}. 
The first is the data split: train and test splits can contain \textit{different levels of annotator disagreement, training scale, and demographic coverage}. 
The second is the modeling framework: demographic information can be incorporated through sociodemographic prompting~\cite{orlikowski2025beyond,sun2025sociodemographic}, learnable special tokens~\cite{lee2024exploring}, demographic embeddings~\cite{sanghani-etal-2025-mcmaster,gordon2022jury}, or more structured mechanisms. 
As a result, demographic information may improve performance in one setting but behave as noise in another. 
Rather than asking whether demographics are useful in general, we study the data and modeling conditions under which they improve performance.

\noindent \textbf{In this work.} We characterize \textit{demographic-gain-favorable regimes}: conditions under which demographic information is more likely to be helpful. 
On the data side, gains are associated with four split properties: (1) less ambiguity in training; (2) more ambiguity in testing; (3) more training data; and (4) more train--test demographic overlap. 
For example, Figure~\ref{fig:demographic-features-help} illustrates a ``reward zone'', where demographics are most likely to help because training disagreement is low but test disagreement is high.
On the modeling side, gains depend on how demographics are incorporated. 
\textit{Separately and selectively} modeling demographic effects is more reliable than directly appending demographics to the input. 
Table~\ref{tab:demo_split_framework_summary} illustrates this interaction: the same dataset can lead to different conclusions under different splits and frameworks.
Motivated by these findings, we propose a gated demographic residual model that treats demographics as a controlled adjustment to text-only predictions. 
We further show that demographic residuals help most on high disagreement or low-confidence examples, can hurt on unambiguous examples, and capture meaningful demographic signals beyond global bias. Our contributions are as follows:
\begin{itemize}
    
    \item We systematically characterize when demographics help in perspective-aware hate speech detection. 
    By analyzing data splits with split-level statistics, such as annotator disagreement, 
    we identify demographic-favorable regimes: (1) less ambiguity in training; (2) more ambiguity in testing; (3) more training data; and (4) more demographic overlap between training and testing.

    \item We show that demographic gains also depend on the modeling framework. 
    Across zero-shot prompting, annotator modeling, finetuned LLMs, and finetuned PLMs, simply providing demographic attributes does not guarantee improvement; the way demographic information is incorporated substantially affects whether it helps or behaves as noise.

    \item We introduce a gated demographic residual model motivated by these regimes. 
    It improves performance especially on high-disagreement and low-confidence examples, and analyses show that it learns meaningful demographic signals rather than global bias.

\end{itemize}


\begin{figure*}[t]
\centering
\begin{subfigure}[t]{0.48\textwidth}
    \centering
    \includegraphics[width=\linewidth]{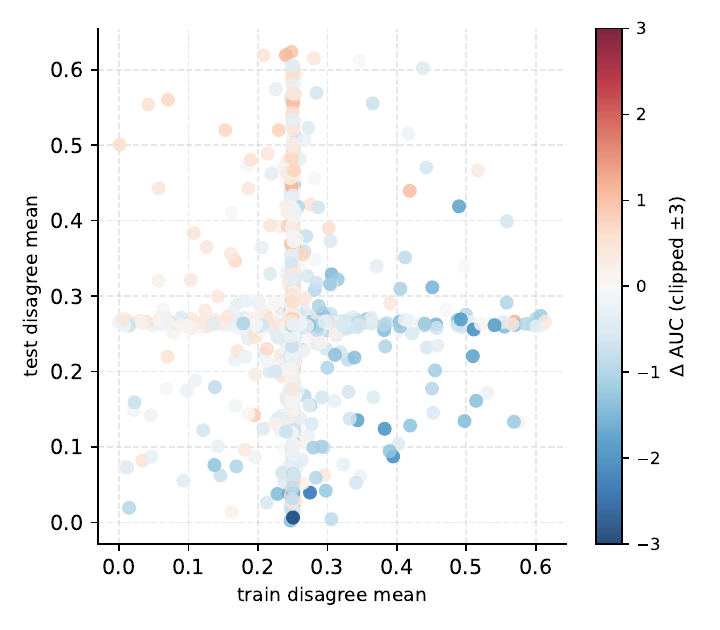}
    \vspace{-7mm}
    \caption{Three-class setting}
    \vspace{-1mm}
    \label{fig:demo_gain_disagree_three_class}
\end{subfigure}
\hfill
\begin{subfigure}[t]{0.48\textwidth}
    \centering
    \includegraphics[width=\linewidth]{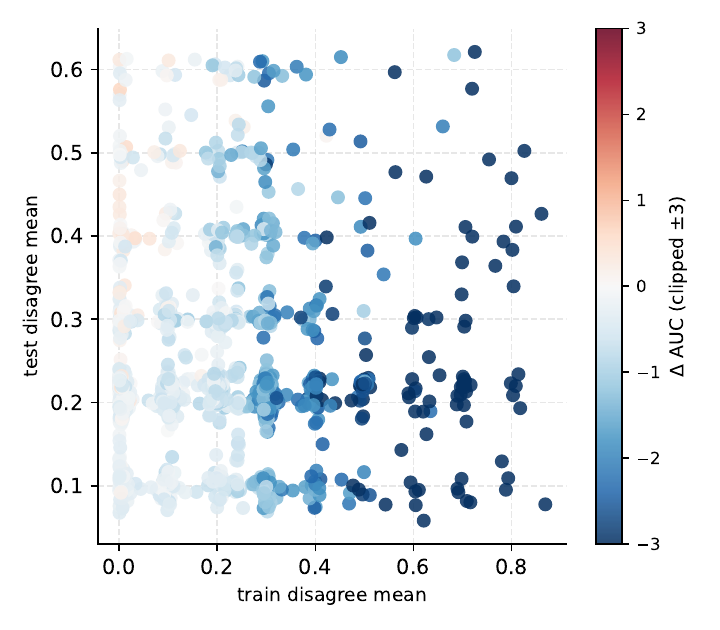} 
    \vspace{-7mm}
    \caption{Binary setting}
    \vspace{-1mm}
    \label{fig:demo_gain_disagree_binary}
\end{subfigure}

\caption{
Effect of train--test disagreement patterns on the AUC gain from demographic features.
Positive gains concentrate in splits with low training disagreement and high test disagreement, especially in the three-class setting. 
In the binary setting, the pattern is weaker and is driven more by low training disagreement.
For the binary setting, disagreement is computed from the original three-class annotations before removing the unclear label.
}
\label{fig:demo_gain_disagreement}
\vspace{-4mm}
\end{figure*}


\vspace{-2mm}
\section{Data Regimes for Demographic Gains}
\label{sec:regimes_split}

Table~\ref{tab:demo_split_framework_summary} suggests that demographic information does not provide a uniform benefit across data splits. 
We therefore study \textit{whether demographic gains vary systematically with properties of the split}. 
We use \textit{demographic-favorable regime} to refer to a region in the space of split-level properties, such as training disagreement, test disagreement, training scale, and demographic coverage, where adding demographic features tends to improve over a text-only model. 
This notion is descriptive rather than a hard threshold rule: it characterizes when demographic information has room to help.

\paragraph{Demographic gain.}
To isolate the effect of data splits, we use a lightweight diagnostic classifier. 
Text is encoded with a frozen Sentence-BERT encoder~\cite{reimers-2019-sentence-bert}. 
For the demographic-aware variant, each demographic attribute is represented by a learnable embedding. 
We mean-pool these embeddings, concatenate the resulting demographic representation with the sentence embedding, and feed the combined representation into an MLP classifier. 
The text-only baseline uses the same classifier without the demographic representation. 
For each split, we define demographic gain as
\[
\Delta \mathrm{AUC} = \mathrm{AUC}_{\text{text+demo}} - \mathrm{AUC}_{\text{text}}.
\]
Positive values indicate that demographic features improve over the corresponding text-only model under the same split.

\begin{table*}[t]
\centering
{\fontsize{10pt}{12pt}\selectfont
\setlength{\tabcolsep}{6pt}
\begin{tabular}{llcccc}
\toprule
\multirow{2}{*}{Regime} 
& \multirow{2}{*}{Measure}
& \multicolumn{2}{c}{MHS 3-class} 
& \multicolumn{2}{c}{MHS 2-class} \\
\cmidrule(lr){3-4} \cmidrule(lr){5-6}
& & $r$ & $\rho$ & $r$ & $\rho$ \\
\midrule

\multirowcell{4}[0pt][l]{1. Less ambiguity in training,\\ \phantom{2. }more ambiguity in testing}
& Train high-disagreement frac.
& $-0.318^{*}$ & $-0.287^{*}$ 
& $-0.885^{*}$ & $-0.851^{*}$ \\
& Mean train disagreement
& $-0.315^{*}$ & $-0.292^{*}$ 
& $-0.885^{*}$ & $-0.851^{*}$ \\
& Mean test disagreement
& $0.259^{*}$ & $0.200^{*}$ 
& $0.012$ & $0.040$ \\
& Test high-disagreement frac.
& $0.259^{*}$ & $0.189^{*}$ 
& $0.017$ & $0.045$ \\

\midrule

\multirow{1}{*}{\makecell[l]{2. More granular measurements}}

& Test uncertain-label frac.
& $0.258^{*}$ & $0.183^{*}$ 
& -- & -- \\

\midrule

\multirow{2}{*}{3. More training data}
& Training unique comments
& $0.244^{*}$ & $0.134^{*}$ 
& $0.702^{*}$ & $0.206^{*}$ \\
& Training records
& $0.201^{*}$ & $0.156^{*}$ 
& $0.526^{*}$ & $0.251^{*}$ \\

\midrule

4. More demographic overlap
& Test demographic overlap
& $0.156^{*}$ & $0.100^{*}$ 
& $0.081^{*}$ & $0.028$ \\

\bottomrule
\end{tabular}
}

\caption{
Correlations between split characteristics and $\Delta$AUC from adding demographic features under the MHS 3-class and 2-class settings.
Positive correlations indicate that larger values of the measure are associated with larger demographic gains; negative correlations indicate the opposite.
Asterisks denote statistically significant correlations.
}
\vspace{-3mm}
\label{tab:demo_gain_correlations}
\end{table*}

\paragraph{Representing each split.}
We use the MHS dataset~\cite{sachdeva2022measuring}, which contains multiple annotations per comment and annotator-level demographic information. 
We evaluate both the original three-class setting, with labels for non-hateful, unclear, and hateful content, and a binary setting that removes the unclear class.
For each setting, we generate 1,400 random train/validation/test splits while holding out both comments and annotator IDs from validation and test. 
For every split, we train the text-only and demographic-aware diagnostic models with five random seeds and use mean AUC to compute $\Delta\mathrm{AUC}$. 
We characterize each split along three dimensions:
\vspace{-1mm}
\begin{enumerate}
    \item \textbf{Ambiguity level:} 
    \begin{itemize}[leftmargin=*]
        \item mean comment-level disagreement (\textit{disagree mean} in Figure~\ref{fig:delta_auc_train_disagree}, ~\ref{fig:demo_gain_disagree_binary} and ~\ref{fig:demo_gain_disagree_three_class}). For a comment $i$, disagreement is computed as its normalized annotation entropy $\frac{-\sum_{c=1}^{C} p_{i,c}\log p_{i,c}}{\log C}$, where $C$ is the number of classes and the higher values indicate greater divergence;
        \item fraction of high disagreement comments (\textit{HD frac}), where we regard a comment with disagree $> 0.6$ as high disagree comment;
        \item fraction of uncertain label (\textit{label-1 frac} in Figure~\ref{fig:delta_auc_test_label_frac}), i.e., fraction of uncertain labels;
    \end{itemize}

    \item \textbf{Training scale:}
    \begin{itemize}[leftmargin=*]
        \item number of annotation records;
        \item number of unique comments (e.g., \textit{train comments} in Figure~\ref{fig:delta_auc_train_size});
    \end{itemize}
    
    \item \textbf{Demographic coverage:}
    \begin{itemize}[leftmargin=*]
        \item test demographic overlap, i.e., the number of test demographic combinations also observed in training (\textit{test demo overlap} in Figure~\ref{fig:delta_auc_demo_overlap}).
        Each demographic combination is a tuple of available annotator attributes.
    \end{itemize}
\end{enumerate}
We conduct this analysis in both the original three-class setting and the binary setting, which allows us to examine how the granularity of ambiguity affects demographic gains. 
We visualize the relation between train and test disagreement in Figure~\ref{fig:demo_gain_disagreement}, and summarize correlations with individual split descriptors in Table~\ref{tab:demo_gain_correlations}. 
Additional univariate plots for these descriptors are provided in Appendix~\ref{app:split_descriptor_plots}. 
We further confirm these results with a multivariate regression in Appendix~\ref{app:multivariate-split-regression}.
These results reveal four demographic-favorable regimes.

\paragraph{Regime 1: lower disagreement training data, higher disagreement testing data.} 
Figure~\ref{fig:demo_gain_disagreement} shows that positive $\Delta\mathrm{AUC}$ values concentrate where training disagreement is low but test disagreement is high. 
This pattern appears in both the three-class and binary settings, although it is stronger in the three-class setting. 
Table~\ref{tab:demo_gain_correlations} supports the same trend: train disagreement is negatively correlated with $\Delta\mathrm{AUC}$ ($r=-0.315$ and $-0.885$), while test disagreement is positively correlated with $\Delta\mathrm{AUC}$ in the three-class setting ($r=0.259$). 
Intuitively, low-disagreement training data provide more stable supervision for learning demographic-conditioned patterns, while high-disagreement test data create cases where text alone is less sufficient. 
Demographics are useful in such cases only if they help explain which annotators give which labels, rather than merely adding another input feature. 
As a negative control, shuffled demographics hurt performance in Appendix~\ref{app:intro_case}, showing that gains require the real correspondence between annotator demographics and labels.


\paragraph{Regime 2: more granular measurement of ambiguity.} 
The role of test ambiguity changes after converting MHS to a binary task. 
In Table~\ref{tab:demo_gain_correlations}, test disagreement is strongly associated with $\Delta\mathrm{AUC}$ in the three-class setting, but becomes weak in the binary setting ($r=0.012$). 
This occurs because removing the intermediate class compresses the disagreement signal: the median binary disagreement is only 7.7\% of the original three-class disagreement. 
This suggests that much of the subjective variation in MHS is captured by the unclear class.
After this class is removed, test disagreement mainly reflects residual binary conflict rather than the broader ambiguity that demographics can help explain. 
Consistently, Figure~\ref{fig:delta_auc_test_label_frac} shows that a higher fraction of intermediate labels is associated with high test disagreement and larger demographic gains. 
This suggests that \textit{finer-grained measurements of subjective perception and ambiguity increase the usefulness of demographic information.}

\paragraph{Regime 3: more training data.} 
More training data expands both textual coverage and demographic coverage. 
Whether this increases demographic gain depends on the relative benefit of the two: additional data may help the text-only model by covering more textual patterns, but it may also help the demographic-aware model by exposing more stable text-demographic associations. 
Table~\ref{tab:demo_gain_correlations} shows positive correlations between $\Delta\mathrm{AUC}$ and both the number of unique training comments ($r=0.244, 0.702$) and the number of annotation records ($r=0.201, 0.526$). 
In our setting, \textit{larger training sets help the model learn demographic patterns beyond what text alone captures.}


\paragraph{Regime 4: more demographic overlap in test set.} 
Demographic information is more useful when the demographic combinations in the test set also appear in the training set. 
If the model has seen a demographic combination during training, it has more evidence about how that group tends to label similar content. 
Table~\ref{tab:demo_gain_correlations} shows that this train--test overlap is positively correlated with $\Delta\mathrm{AUC}$, especially in the three-class setting ($r=0.156$). 
The effect is weaker in the binary setting ($r=0.081$) because removing the unclear class reduces the borderline cases where demographic differences are most informative. 
Thus, \textit{demographic overlap matters most when the task retains enough ambiguity for demographic variation to be useful.}

\paragraph{Implications.}
These regimes turn the question of whether demographics help from a post-hoc observation into a diagnostic question about the data and model.
They also suggest a modeling principle: \textit{demographics should be used selectively rather than applied uniformly to all examples.}
In the next section, we test whether this \textit{regime-aware view} can make demographic-aware modeling more reliable.

\begin{table*}[t]
\centering
\scriptsize
\setlength{\tabcolsep}{3.5pt}
\renewcommand{\arraystretch}{1.08}
\begin{tabular}{p{2.85cm}p{1.35cm}rrrrrr rrrrrr}
\toprule
\multirow{2}{*}{Backbone}
& \multirow{2}{*}{Input}
& \multicolumn{6}{c}{MHS}
& \multicolumn{6}{c}{POPQUORN} \\
\cmidrule(lr){3-8} \cmidrule(lr){9-14}
& & Acc. & Prec. & Rec. & F1 & AUC & $\Delta$AUC
  & Acc. & Prec. & Rec. & F1 & AUC & $\Delta$AUC \\
\midrule

\group{Zero-shot} \\
LLaMA-2-7B-chat-hf & Text
& 40.97 & 64.98 & 58.12 & 39.39 & 73.02 & \na
& 32.35 & 56.10 & 53.20 & 31.08 & 60.71 & \na \\
\rowcolor{demoRow}
LLaMA-2-7B-chat-hf & +Demo
& 69.04 & 62.15 & 61.97 & 62.06 & 70.78 & \drop{-2.24}
& 27.43 & 51.62 & 50.43 & 24.65 & 53.81 & \drop{-6.90} \\

Mistral-7B-v0.3 & Text
& 33.04 & 55.32 & 51.61 & 29.56 & 62.15 & \na
& 45.00 & 53.12 & 53.83 & 44.41 & 54.69 & \na \\
\rowcolor{demoRow}
Mistral-7B-v0.3 & +Demo
& 29.04 & 55.09 & 50.05 & 22.67 & 63.93 & \gain{1.78}
& 75.42 & 47.78 & 49.97 & 43.20 & 46.61 & \drop{-8.08} \\

Qwen2-7B & Text
& 40.06 & 64.48 & 57.42 & 38.29 & 77.01 & \na
& 53.78 & 58.53 & \textbf{61.19} & 52.44 & 66.14 & \na \\
\rowcolor{demoRow}
Qwen2-7B & +Demo
& 40.47 & 64.56 & 57.70 & 38.80 & 76.23 & \drop{-0.78}
& 52.11 & 56.88 & 59.03 & 50.74 & 63.91 & \drop{-2.23} \\

\midrule
\group{Annotator modeling} \\
TID-DeBERTa-v3-base & Text
& 80.59 & 76.54 & 76.45 & 76.39 & 86.82 & \na
& 75.51 & 64.37 & 56.19 & 55.67 & 71.49 & \na \\
\rowcolor{demoRow}
TID-DeBERTa-v3-base & +Demo
& 80.48 & 76.33 & 77.85 & 76.96 & 87.00 & \gain{0.18}
& 75.78 & 64.87 & 56.42 & 56.33 & 72.30 & \gain{0.81} \\

\midrule
\group{Finetuned LLMs} \\
LLaMA-3.2-1B-SPT & Text+Demo
& 79.55 & 75.24 & 74.15 & 74.60 & 84.50 & \na
& 74.75 & 51.76 & 52.47 & 48.21 & 64.72 & \na \\

Qwen2.5-0.5B-LoRA & Text
& 80.23 & 76.12 & 74.48 & 75.20 & 85.50 & \na
& 71.14 & 56.90 & 55.54 & 55.81 & 61.44 & \na \\
\rowcolor{demoRow}
Qwen2.5-0.5B-LoRA & +Demo
& 80.76 & 76.86 & 75.01 & 75.80 & 86.10 & \gain{0.60}
& 74.78 & 63.25 & 58.08 & 58.74 & 68.97 & \gain{7.53} \\

LLaMA-3.2-1B-LoRA & Text
& 81.17 & 77.34 & 75.67 & 76.40 & 87.05 & \na
& 73.64 & 59.73 & 55.36 & 55.29 & 66.18 & \na \\
\rowcolor{demoRow}
LLaMA-3.2-1B-LoRA & +Demo
& \textbf{81.83} & \textbf{78.28} & 76.17 & \textit{77.08} & \textbf{87.54} & \gain{0.49}
& 75.18 & 63.63 & 57.88 & 58.47 & 68.61 & \gain{2.43} \\

\midrule
\group{Finetuned PLMs} \\
ToxDect-RoBERTa & Text
& 80.41 & 76.47 & 76.56 & 76.30 & 86.86 & \na
& 74.32 & 62.21 & 56.53 & 56.49 & 71.68 & \na \\
\rowcolor{demoRow}
ToxDect-RoBERTa & +Demo
& 80.79 & 76.79 & 76.01 & 76.31 & 86.75 & \drop{-0.11}
& 74.38 & 62.68 & 57.78 & 58.02 & 71.46 & \drop{-0.22} \\
\rowcolor{oursRow}
ToxDect-RoBERTa & \textbf{Ours}
& 80.93 & 76.81 & 77.21 & 76.99 & 87.02 & \textbf{\gainx{0.16}{1.45}}
& \textit{76.12} & \textit{65.96} & 59.20 & \textit{59.98} & 73.73 & \textbf{\gainx{2.27}{10.3}} \\

BERTweet & Text
& 80.80 & 76.80 & 76.94 & 76.74 & 87.07 & \na
& 74.60 & 61.49 & 55.15 & 54.10 & 71.55 & \na \\
\rowcolor{demoRow}
BERTweet & +Demo
& 80.58 & 76.53 & \textbf{77.95} & 77.05 & 87.22 & \gain{0.15}
& 75.89 & 65.92 & 58.48 & 58.81 & \textit{73.94} & \gain{2.39} \\
\rowcolor{oursRow}
BERTweet & \textbf{Ours}
& \textit{81.58} & \textit{77.71} & \textit{77.70} & \textbf{77.31} & \textit{87.39} & \gainx{0.32}{2.13}
& \textbf{76.80} & \textbf{67.51} & \textit{59.28} & \textbf{60.19} & \textbf{75.71} & \gainx{4.16}{1.74} \\

\bottomrule
\end{tabular}
\caption{
Main performance comparison on MHS and POPQUORN.
$\Delta$AUC is computed relative to the corresponding text-only baseline with the same backbone and dataset; for Demo-TID, the baseline is TID.
Shaded +Demo rows indicate demographic information, and blue rows denote our residual method.
For our method, bold parenthesized values indicate cases where ordinary +Demo decreases AUC, but our method reverses the effect and yields a positive $\Delta$AUC.
Bold and italic mark the best and second-best results within each dataset and metric.
}
\vspace{-3mm}
\label{tab:main_results_compact}
\end{table*}

\section{Regime-Aware Gated Demographic Residual Model}
\label{sec:residual_model}
Regime 1 shows that disagreement plays different roles in training and testing:
high training disagreement can make demographic-label associations noisy, while high test disagreement creates opportunities for demographic information to help when text alone is insufficient. 
Motivated by this asymmetry, we use demographics as a controlled residual correction to a text-only classifier. 
The text model provides the primary prediction, and demographic information adjusts the logits mainly when the text-only prediction is uncertain.

\subsection{Gated Demographic Residual Adapter}
Given text $x$ and annotator demographics $d$, the text encoder produces a representation $h$ and text-only logits $\mathbf{z}^{\text{text}}$. 
We encode demographics with learnable embeddings, mean-pool them into $\mathbf{e}(d)$, and compute a residual correction $\mathbf{r}=g_{\phi}([h;\mathbf{e}(d)])$, where $[\cdot;\cdot]$ denotes concatenation.

To make the correction selective, we gate it by text-only uncertainty.
Let $\mathbf{p}^{\text{text}}=\mathrm{softmax}(\mathbf{z}^{\text{text}})$ and compute normalized entropy
\begin{equation}
    u =
\frac{
    -\sum_{k=1}^{K}
    p^{\text{text}}_k
    \log p^{\text{text}}_k
}{\log K},
\end{equation}
where $K$ is the number of classes. 
The gate and final logits are
\begin{equation}
\alpha = \sigma\!\left(\frac{u-\tau}{T}\right),
\qquad
\mathbf{z}=\mathbf{z}^{\text{text}}+\alpha\mathbf{r},
\end{equation}
where $\tau$ controls when the residual becomes active and $T$ controls transition sharpness. 
Thus, demographics have limited influence when the text model is confident, but can adjust predictions when the text signal is uncertain.

\subsection{Training Objective}
We first train a text-only classifier with annotation-level labels and comment-level soft labels.
Let $\mathbf{p}^{\text{text}}=\mathrm{softmax}(\mathbf{z}^{\text{text}})$, and let $\mathbf{q}_{c(i)}$ denote the empirical label distribution of the comment associated with annotation $i$:
\begin{equation}
\begin{aligned}
\mathcal{L}_{\text{text}}
=&\  \mathrm{CE}(y_i, \mathbf{p}^{\text{text}}_i) \\
&+ \lambda_s \mathrm{CE}(\mathbf{q}_{c(i)}, \mathbf{p}^{\text{text}}_{c(i)}).
\end{aligned}
\end{equation}
The hard-label term supports annotation-level prediction, while the soft-label term encourages uncertainty to reflect aggregate annotator disagreement.

We then freeze the text-only model and train the demographic residual adapter with a gate-weighted cross-entropy loss:
\vspace{-1mm}
\begin{equation}
\mathcal{L}_{\text{res}}
=
\frac{1}{N}\sum_{i=1}^{N}
\left((1-\rho)+\rho\alpha_i\right)
\mathrm{CE}(y_i,\mathbf{p}_i),
\end{equation}
where $\mathbf{p}_i=\mathrm{softmax}(\mathbf{z}_i)$. 
The parameter $\rho \in [0,1]$ controls how strongly uncertain examples are upweighted, encouraging the adapter to learn demographic corrections mainly where text-only evidence is weak.

\begin{figure*}[t]
\centering
\begin{subfigure}[t]{0.48\textwidth}
    \centering
    \includegraphics[width=\linewidth]{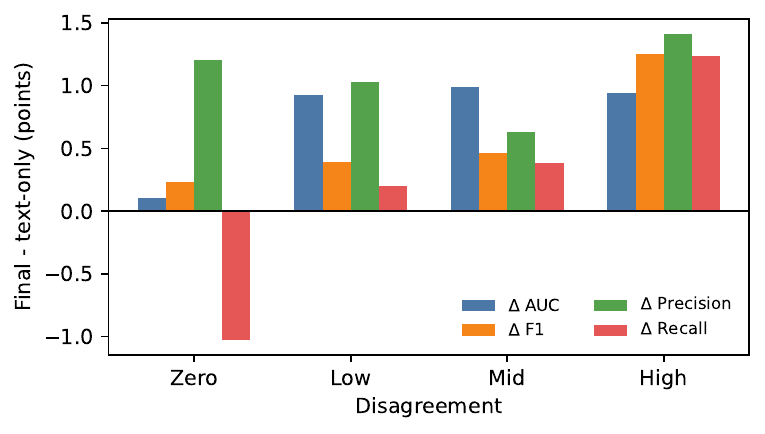}
    \caption{Residual gain by disagreement level}
    \label{fig:resid_gain_disagree_buckets}
\end{subfigure}
\hfill
\begin{subfigure}[t]{0.48\textwidth}
    \centering
    \includegraphics[width=\linewidth]{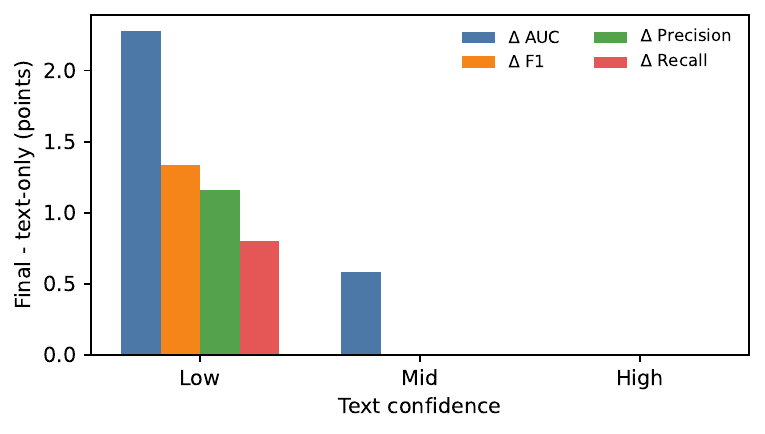}
    \caption{Residual gain by text confidence level}
    \label{fig:resid_gain_confidence_buckets}
\end{subfigure}

\caption{
Demographic residual gains across test subsets.
Bars show the change from the text-only model to the residual model. 
Disagreement groups use comment-level disagreement, with zero-disagreement comments separated, and the rest split into equal-frequency low, medium, and high bins. 
Text confidence is measured by the text-only decision-boundary margin \(c_i=|p_i^{\text{text}}(y=1\mid x_i)-0.5|\), where smaller values indicate lower confidence.
}
\label{fig:resid_gain}
\vspace{-4mm}
\end{figure*}

\section{Modeling and Diagnosing Demographic Gains}
We evaluate two questions. 
First, how do different modeling frameworks affect demographic gains? 
Second, does our residual model use demographic information selectively and meaningfully, as suggested by the regimes in Section~\ref{sec:regimes_split}?

\paragraph{Dataset.}
We evaluate on MHS~\cite{sachdeva2022measuring} and POPQUORN~\cite{pei2023annotator}. 
(1) MHS contains 50,070 comments and 11,143 annotators, with an average of 2.7 annotations per comment. 
For the binary setup, we remove the intermediate class and construct train, validation, and test splits that are disjoint in both comments and annotators, yielding 72,130/7,558/7,596 annotation instances.
(2) POPQUORN contains 1,500 comments, 262 annotators, and an average of 8.7 annotations per comment. 
We use its offensiveness subset and binarize the original five-point ratings by mapping ratings 1--2 to non-offensive and ratings 3--5 to offensive. 
Because each POPQUORN annotator labels many comments, enforcing unseen-annotator splits would leave too few annotation instances for reliable training and evaluation. 
We therefore hold out unseen comments only, yielding 9,098/1,969/1,969 annotation instances for train, validation, and test.
Additional dataset details are provided in Appendix~\ref{app:dataset}.

\paragraph{Model frameworks.}
We compare four families of demographic-aware frameworks: 
(1) \textit{Zero-shot prompting}, using LLaMA-2-7B-Chat~\cite{touvron2023llama}, Mistral-7B-v0.3~\cite{Jiang2023Mistral7}, and Qwen2-7B~\cite{Yang2024Qwen2TR}. 
(2) \textit{Annotator modeling}, using TID~\cite{deng2023you}; for the TID setting, we use random annotator IDs under unseen-annotator splits, and for Demo-TID we replace annotator IDs with demographic attributes. 
(3) \textit{Finetuned LLMs}, including LLaMA-3.2-1B-SPT with soft prompt tuning~\cite{lester2021power}, and LLaMA-3.2-1B~\cite{grattafiori2024llama} and Qwen2.5-0.5B~\cite{Yang2024Qwen25TR} with LoRA~\cite{hu2022lora}; demographic information is incorporated through sociodemographic prompts. 
(4) \textit{Finetuned PLMs}, using BERTweet~\cite{nguyen2020bertweet} and ToxDect-RoBERTa~\cite{zhou2020challenges}; for the +Demo baseline, we prepend learnable demographic tokens to the text, following prior work~\cite{lee2023exploring,fleisig2023majority}. 
Prompt templates and details are provided in Appendix~\ref{app:exp_details}.

\subsection{How Modeling Frameworks Shape Demographic Gains}
\label{sec:regimes_model}

Table~\ref{tab:main_results_compact} shows that demographic gains vary substantially across modeling frameworks and datasets. 
First, gains are generally larger on POPQUORN than on MHS. 
POPQUORN contains only 1,050 training comments but many annotations per comment, and it was intentionally sampled to contain high-disagreement examples. 
This weakens text-only models, especially finetuned LLMs, and leaves more room for demographic information to explain annotator variation.
Second, \textit{naive demographic injection is unreliable}. 
Zero-shot demographic prompting yields inconsistent or negative gains, even though LLMs may encode broad social priors. 
Among finetuned models, the effect depends on how demographics are represented. 
For PLMs, learned demographic tokens must acquire their meaning from task supervision alone, so the demographic signal can be overwhelmed by strong textual cues, especially on MHS. 
In contrast, LoRA-finetuned LLMs benefit more from sociodemographic prompts, likely because demographic descriptions are expressed in natural language and can reuse pretrained semantic knowledge.
Third, \textit{more structured demographic conditioning is more stable}. 
Demo-TID yields mild gains on both datasets, and our residual model further improves over ordinary demographic-token finetuning. 
Notably, it turns the negative +Demo effect of ToxDect-RoBERTa into a positive gain. 
These results suggest that demographic information is most reliable when its effect is controlled rather than directly injected into the input.

\subsection{Where Does the Demographic Residual Help or Hurt?}
\label{sec:resid_gain}

We next examine whether the residual model behaves consistently with Regime~1.
Rather than only asking whether demographics improve overall performance, we ask where the gain comes from. 
Figure~\ref{fig:resid_gain} shows that the residual is not uniformly beneficial. 
Its gains concentrate on high-disagreement and low-confidence examples, where the text-only model is less reliable and demographic context has more room to affect the decision. 
By contrast, the residual can hurt on zero-disagreement examples, where the label signal is already clear and demographic adjustment may introduce unnecessary variation. 
This supports the central intuition of Regime~1: demographic information helps most under ambiguous test conditions, but can be harmful when the text signal is already decisive.



\begin{figure*}[t]
  \centering
  \begin{minipage}[t]{0.48\textwidth}
    \centering
    \includegraphics[width=\linewidth]{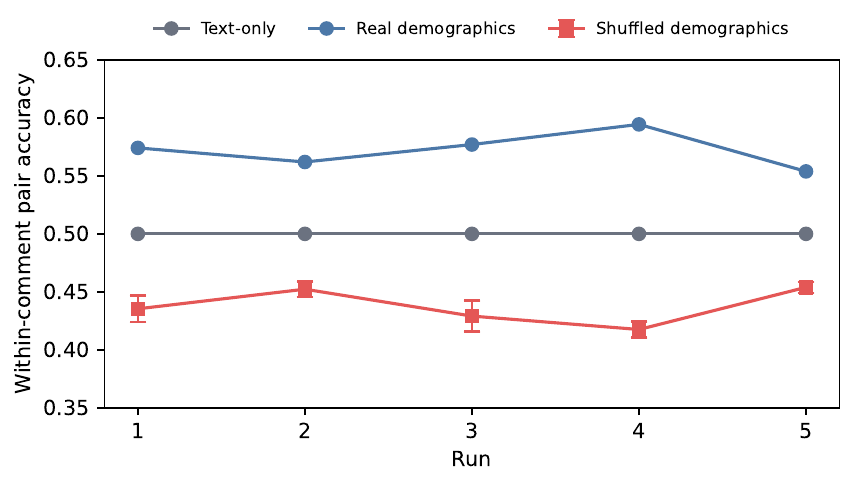}
    \caption{Real vs. shuffled demographics in within-comment pairwise accuracy.}
    \label{fig:within_com_acc}
    \vspace{-3mm}
  \end{minipage}
  \hfill
  \begin{minipage}[t]{0.48\textwidth}
    \centering
    \includegraphics[width=\linewidth]{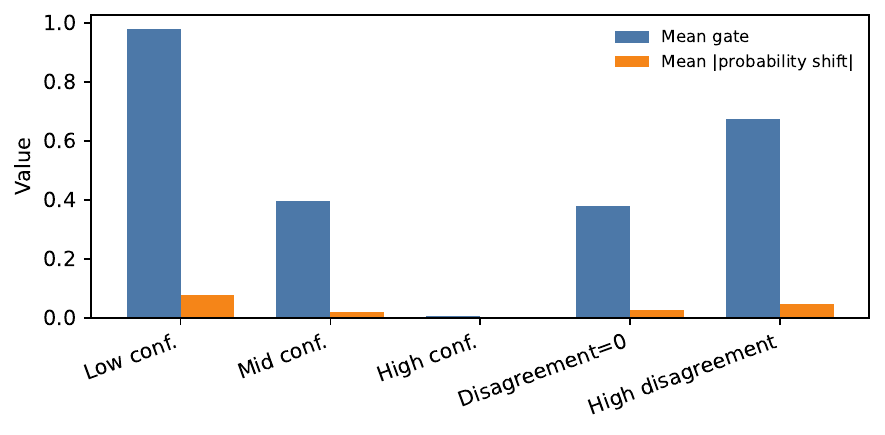}
    \caption{Gate selectivity. Higher gate values indicate stronger demographic residual adjustment.}
    \label{fig:gate_select}
    \vspace{-3mm}
  \end{minipage}
\end{figure*}

\subsection{Does the Residual Capture Demographic Signal?}
\label{sec:within_com_acc}
We test whether the residual model uses demographics as meaningful conditioning variables, rather than relying on global demographic priors. 
For each disagreed comment in the binary setting, we form pairs of annotators with opposite labels. 
A pair is correct if the model assigns a higher positive-class probability to the annotator who labeled the comment positive. 
This within-comment comparison controls for textual variation and directly tests whether demographics help explain different reactions to the same comment. 
Figure~\ref{fig:within_com_acc} shows that real demographics consistently outperform both the text-only and shuffled-demographic baseline.  
The gap between real and shuffled demographics suggests that the residual model learns useful demographic signals beyond global demographic frequencies or comment-level bias.

\subsection{Selective Use of Demographics}
\label{sec:gate_select}

We examine whether the gate applies demographic adjustments selectively rather than uniformly. 
Figure~\ref{fig:gate_select} shows that gate values are higher when the text-only model is less confident, and lower when the text-only prediction is more decisive. 
The same pattern appears across disagreement groups: higher disagreement examples receive stronger demographic adjustment. 
This behavior matches our regime view, where demographic information should influence predictions mainly when the text signal is uncertain or ambiguous.

\section{Related Work}

Our results provide a regime-based view of prior work on disagreement, demographic annotation, and demographic-aware modeling. 
Most studies do not directly ask when demographics help prediction. 
Instead, they often focus on one favorable condition, or use settings that implicitly satisfy part of the regimes identified in this paper. 
We make these conditions explicit and show how they jointly shape demographic gains.

\paragraph{Disagreement in training and testing.}
Some studies model soft label distributions~\cite{leonardelli2023semeval}, while others model annotator-specific behavior with annotator-level heads or representations~\cite{davani2022dealing}. 
These methods implicitly address the negative effect of high-disagreement training data. 
Other work highlights demographic effects on divergent or borderline examples~\cite{yin2023annobert,pei2023annotator}, and persona-based simulation is most useful when examples exhibit disagreement~\cite{hu2024quantifying}. 
These findings align with our finding that train and test disagreement play different roles: demographics help most when training disagreement is low but test disagreement is high.

\paragraph{Granularity of ambiguity.}
Existing studies measure hate speech with severity levels~\cite{ross2017measuring}, continuous scores~\cite{kennedy2020constructing}, or multiple components such as insult, and dehumanization~\cite{sachdeva2022measuring}. 
These designs preserve borderline judgments, consistent with our finding that demographics become less predictive when coarse-grained labeling obscures ambiguity.

\paragraph{Training scale and demographic coverage.}
Prior work finds that demographic features help more when groups are diverse, balanced, and well represented~\cite{tahaeibeyond}. 
Frequent demographic combinations in training are also easier for models to use effectively~\cite{orlikowski2025beyond}. 
These align with our regimes on training scale and demographic overlap: more examples and greater train--test demographic overlap make demographic-label associations easier to learn.

\paragraph{Modeling framework.}
Different demographic-aware frameworks can produce different outcomes~\cite{zhang2026modeling}, and zero-shot sociodemographic prompting often yields inconsistent gains~\cite{sun2025sociodemographic,beck2024sensitivity,lutz2025prompt}. 
This agrees with our conclusion that demographics are not automatically useful once provided to the model; their effect depends on the framework used to condition on them.

\section{Conclusion}

This paper reframes the usefulness of demographic information as a regime-dependent question in perspective-aware hate speech detection. 
Rather than treating demographics as uniformly beneficial, we show that their usefulness depends on both the data regime and the modeling framework. 
Through split analysis on MHS, we identify demographic-favorable regimes: (1) low training disagreement, high test disagreement, (2) fine-grained ambiguity measurement, (3) sufficient training data, and (4) greater train-test demographic overlap. 
These findings help explain why demographic-aware models often produce inconsistent gains across datasets and experimental settings.

The data-regime analysis also motivates how demographics should be modeled. 
Since high training disagreement can make demographic-label associations noisy, while high test ambiguity creates cases where text alone is insufficient, demographics should not be applied uniformly to all examples. 
Motivated by this insight, we introduce a gated demographic residual model that uses text-only predictions as the primary signal and allows demographic information to adjust predictions mainly when the text signal is uncertain. 
Results on MHS and POPQUORN show that this design is effective, with gains concentrated on high-disagreement and low-confidence examples. 
Overall, our results show that demographic information is not useful by default; it becomes useful only under favorable data and modeling regimes.

\section{Limitations}
\label{sec:limitation}

Our analysis is limited to two English hate speech and offensiveness datasets, MHS and POPQUORN. 
Although these datasets are well suited for studying annotator disagreement and demographic information, the identified regimes may not transfer directly to other domains or demographic schemas. 
In particular, POPQUORN is intentionally enriched with high disagreement examples and contains fewer unique comments, so it is less suitable for discovering split-level regimes from scratch.

The demographic attributes available in existing datasets are also coarse and incomplete. 
They cannot capture the full complexity of identity, or within-group variation. 
Therefore, the demographic effects studied here should be interpreted as dataset-specific annotation patterns rather than general claims about social groups.

Our regime analysis is descriptive rather than causal. 
The correlations between split properties and demographic gain indicate when demographics are more likely to help, but they do not prove that each factor independently causes the gain. 
These regimes are intended for dataset diagnosis and evaluation design, rather than as direct inference-time features. 
Some descriptors, such as test disagreement, rely on aggregate annotations and are not assumed to be available at deployment time.

Another limitation is that the regime factors are not fully independent. 
For example, test disagreement may covary with uncertain-label frequency, label distribution, or training size. 
Although our analysis identifies consistent associations between these factors and demographic gains, it does not fully disentangle their individual effects.

Finally, our split analysis and residual model use a simple mean-pooled demographic representation. 
We choose this naive representation in the regime analysis because it provides a conservative setting for observing when demographic information can help; stronger demographic encoders may yield larger gains, but they still need favorable data conditions to make demographic information useful.
In the residual model, mean pooling is also not intended as the optimal way to model complex demographic interactions. 
Our goal is to test a regime-aware modeling principle, rather than to fully solve demographic representation learning.

\section{Ethical Considerations}
This work studies how demographic information affects hate speech and offensiveness classification. Demographics are sensitive, and using them in predictive models can introduce privacy risks, reinforce stereotypes, or amplify spurious associations between social groups and labels. 
Our goal is not to profile demographic groups or claim that any group has a fixed perception of offensiveness. Instead, we use demographics to analyze annotator disagreement and to study when demographic conditioning helps or harms model predictions. 
Although our residual model is designed to use demographics selectively, this does not eliminate ethical risks. A model may still learn correlations that reflect dataset artifacts, sampling bias, or stereotypes. For this reason, we treat the model as an analytical tool rather than a deployable moderation system. 
In real-world applications, demographic-aware models should only be used with informed consent, and strong privacy protections.

We use only existing public research datasets and do not collect new personal information. 
The datasets provide annotator demographics in aggregated or categorical form rather than direct identifiers such as names or contact information. 
Because the datasets contain hateful or offensive text, we avoid reproducing unnecessary examples and report only aggregate analyses.

Our experiments rely on existing annotated datasets, whose demographic categories are necessarily coarse and incomplete. These categories cannot capture the full complexity of identity or within-group variation. We therefore interpret demographic effects as dataset-specific annotation patterns rather than general claims about social groups. 
All the models employed are publicly accessible and their use is consistent with their intended purposes. 
All model training and inference were conducted using an NVIDIA Quadro RTX 6000 GPU. The overall scale of the residual model depends on the choice of the backbone, while the residual component introduces only approximately 0.385M parameters.

\section{Use of AI Assistants}
We used AI language assistants, such as ChatGPT, to improve writing clarity and grammar, and assist with code debugging. 
We used an AI image-generation tool to create the coin in Figure~\ref{fig:demographic-features-help}; the final figure content, layout, and labels were manually reviewed and edited by the authors.
All research ideas, experimental design, analysis, code, and final manuscript content were reviewed and finalized by the human authors.

\bibliography{ref}

@article{lee2023exploring,
  title={Exploring cross-cultural differences in English hate speech annotations: From dataset construction to analysis},
  author={Lee, Nayeon and Jung, Chani and Myung, Junho and Jin, Jiho and Camacho-Collados, Jose and Kim, Juho and Oh, Alice},
  journal={arXiv preprint arXiv:2308.16705},
  year={2023}
}

@article{nguyen2020bertweet,
  title={BERTweet: A pre-trained language model for English Tweets},
  author={Nguyen, Dat Quoc and Vu, Thanh and Nguyen, Anh Tuan},
  journal={arXiv preprint arXiv:2005.10200},
  year={2020}
}

@book{zhou2020challenges,
  title={Challenges in automated debiasing for toxic language detection},
  author={Zhou, Xuhui},
  year={2020},
  publisher={University of Washington}
}

@article{sap2019social,
  title={Social bias frames: Reasoning about social and power implications of language},
  author={Sap, Maarten and Gabriel, Saadia and Qin, Lianhui and Jurafsky, Dan and Smith, Noah A and Choi, Yejin},
  journal={arXiv preprint arXiv:1911.03891},
  year={2019}
}

@inproceedings{sap2022annotators,
  title={Annotators with attitudes: How annotator beliefs and identities bias toxic language detection},
  author={Sap, Maarten and Swayamdipta, Swabha and Vianna, Laura and Zhou, Xuhui and Choi, Yejin and Smith, Noah A},
  booktitle={Proceedings of the 2022 conference of the north american chapter of the association for computational linguistics: Human language technologies},
  pages={5884--5906},
  year={2022}
}

@inproceedings{sachdeva2022measuring,
  title={The measuring hate speech corpus: Leveraging rasch measurement theory for data perspectivism},
  author={Sachdeva, Pratik and Barreto, Renata and Bacon, Geoff and Sahn, Alexander and Von Vacano, Claudia and Kennedy, Chris},
  booktitle={Proceedings of the 1st Workshop on Perspectivist Approaches to NLP@ LREC2022},
  pages={83--94},
  year={2022}
}

@article{lester2021power,
  title={The power of scale for parameter-efficient prompt tuning},
  author={Lester, Brian and Al-Rfou, Rami and Constant, Noah},
  journal={arXiv preprint arXiv:2104.08691},
  year={2021}
}

@inproceedings{deng2023you,
  title={You are what you annotate: Towards better models through annotator representations},
  author={Deng, Naihao and Zhang, Xinliang and Liu, Siyang and Wu, Winston and Wang, Lu and Mihalcea, Rada},
  booktitle={Findings of the Association for Computational Linguistics: EMNLP 2023},
  pages={12475--12498},
  year={2023}
}

@article{fleisig2023majority,
  title={When the majority is wrong: Modeling annotator disagreement for subjective tasks},
  author={Fleisig, Eve and Abebe, Rediet and Klein, Dan},
  journal={arXiv preprint arXiv:2305.06626},
  year={2023}
}

@inproceedings{davani2024d3code,
  title={D3CODE: Disentangling disagreements in data across cultures on offensiveness detection and evaluation},
  author={Davani, Aida and Diaz, Mark and Baker, Dylan and Prabhakaran, Vinodkumar},
  booktitle={Proceedings of the 2024 Conference on Empirical Methods in Natural Language Processing},
  pages={18511--18526},
  year={2024}
}

@inproceedings{orlikowski2023ecological,
  title={The ecological fallacy in annotation: Modeling human label variation goes beyond sociodemographics},
  author={Orlikowski, Matthias and R{\"o}ttger, Paul and Cimiano, Philipp and Hovy, Dirk},
  booktitle={Proceedings of the 61st Annual Meeting of the Association for Computational Linguistics (Volume 2: Short Papers)},
  pages={1017--1029},
  year={2023}
}

@inproceedings{sun2025sociodemographic,
  title={Sociodemographic prompting is not yet an effective approach for simulating subjective judgments with LLMs},
  author={Sun, Huaman and Pei, Jiaxin and Choi, Minje and Jurgens, David},
  booktitle={Proceedings of the 2025 Conference of the Nations of the Americas Chapter of the Association for Computational Linguistics: Human Language Technologies (Volume 2: Short Papers)},
  pages={845--854},
  year={2025}
}

@inproceedings{sorensen2025value,
  title={Value profiles for encoding human variation},
  author={Sorensen, Taylor and Mishra, Pushkar and Patel, Roma and Tessler, Michael Henry and Bakker, Michiel A and Evans, Georgina and Gabriel, Iason and Goodman, Noah and Rieser, Verena},
  booktitle={Proceedings of the 2025 Conference on Empirical Methods in Natural Language Processing},
  pages={2047--2095},
  year={2025}
}

@inproceedings{orlikowski2025beyond,
  title={Beyond demographics: Fine-tuning large language models to predict individuals’ subjective text perceptions},
  author={Orlikowski, Matthias and Pei, Jiaxin and R{\"o}ttger, Paul and Cimiano, Philipp and Jurgens, David and Hovy, Dirk},
  booktitle={Proceedings of the 63rd Annual Meeting of the Association for Computational Linguistics (Volume 1: Long Papers)},
  pages={2092--2111},
  year={2025}
}

@inproceedings{tahaei2025demographic,
  title={Demographic Features for Annotation-Aware Classification},
  author={Tahaei, Narjes and Bergler, Sabine},
  booktitle={Proceedings of the 15th International Conference on Recent Advances in Natural Language Processing-Natural Language Processing in the Generative AI Era},
  pages={1232--1236},
  year={2025}
}

@inproceedings{jaggi2024accurate,
  title={Accurate and data-efficient toxicity prediction when annotators disagree},
  author={Jaggi, Harbani and Murali, Kashyap Coimbatore and Fleisig, Eve and Biyik, Erdem},
  booktitle={Proceedings of the 2024 Conference on Empirical Methods in Natural Language Processing},
  pages={21910--21917},
  year={2024}
}

@inproceedings{gordon2022jury,
  title={Jury learning: Integrating dissenting voices into machine learning models},
  author={Gordon, Mitchell L and Lam, Michelle S and Park, Joon Sung and Patel, Kayur and Hancock, Jeff and Hashimoto, Tatsunori and Bernstein, Michael S},
  booktitle={Proceedings of the 2022 CHI Conference on Human Factors in Computing Systems},
  pages={1--19},
  year={2022}
}

@inproceedings{hu2024quantifying,
  title={Quantifying the persona effect in LLM simulations},
  author={Hu, Tiancheng and Collier, Nigel},
  booktitle={Proceedings of the 62nd Annual Meeting of the Association for Computational Linguistics (Volume 1: Long Papers)},
  pages={10289--10307},
  year={2024}
}

@article{frenda2025perspectivist,
  title={Perspectivist approaches to natural language processing: a survey},
  author={Frenda, Simona and Abercrombie, Gavin and Basile, Valerio and Pedrani, Alessandro and Panizzon, Raffaella and Cignarella, Alessandra Teresa and Marco, Cristina and Bernardi, Davide},
  journal={Language Resources and Evaluation},
  volume={59},
  number={2},
  pages={1719--1746},
  year={2025},
  publisher={Springer}
}

@inproceedings{cabitza2023toward,
  title={Toward a perspectivist turn in ground truthing for predictive computing},
  author={Cabitza, Federico and Campagner, Andrea and Basile, Valerio},
  booktitle={Proceedings of the AAAI Conference on Artificial Intelligence},
  volume={37},
  number={6},
  pages={6860--6868},
  year={2023}
}

@article{xu2025modeling,
  title={Modeling annotator disagreement with demographic-aware experts and synthetic perspectives},
  author={Xu, Yinuo and Derricks, Veronica and Earl, Allison and Jurgens, David},
  journal={arXiv preprint arXiv:2508.02853},
  year={2025}
}

@inproceedings{lee2024exploring,
  title={Exploring cross-cultural differences in English hate speech annotations: From dataset construction to analysis},
  author={Lee, Nayeon and Jung, Chani and Myung, Junho and Jin, Jiho and Camacho-Collados, Jose and Kim, Juho and Oh, Alice},
  booktitle={Proceedings of the 2024 Conference of the North American Chapter of the Association for Computational Linguistics: Human Language Technologies (Volume 1: Long Papers)},
  pages={4205--4224},
  year={2024}
}

@inproceedings{sanghani-etal-2025-mcmaster,
  title = "{M}c{M}aster at {L}e{W}i{D}i-2025: Demographic-Aware {R}o{BERT}a",
  author = "Sanghani, Aadi and Azadi, Sarvin and Jethra, Virendra and Welch, Charles",
  booktitle = "Proceedings of the The 4th Workshop on Perspectivist Approaches to NLP",
  year = "2025",
  address = "Suzhou, China",
  publisher = "Association for Computational Linguistics",
  pages = "208--218",
  url = "https://aclanthology.org/2025.nlperspectives-1.18/",
  doi = "10.18653/v1/2025.nlperspectives-1.18"
}

@inproceedings{yin2023annobert,
  title={Annobert: Effectively representing multiple annotators’ label choices to improve hate speech detection},
  author={Yin, Wenjie and Agarwal, Vibhor and Jiang, Aiqi and Zubiaga, Arkaitz and Sastry, Nishanth},
  booktitle={Proceedings of the International AAAI Conference on Web and Social Media},
  volume={17},
  pages={902--913},
  year={2023}
}

@inproceedings{wan2023everyone,
  title={Everyone’s voice matters: Quantifying annotation disagreement using demographic information},
  author={Wan, Ruyuan and Kim, Jaehyung and Kang, Dongyeop},
  booktitle={Proceedings of the AAAI Conference on Artificial Intelligence},
  volume={37},
  number={12},
  pages={14523--14530},
  year={2023}
}

@inproceedings{reimers-2019-sentence-bert,
  title = "Sentence-{BERT}: Sentence Embeddings using Siamese {BERT}-Networks",
  author = "Reimers, Nils and Gurevych, Iryna",
  booktitle = "Proceedings of the 2019 Conference on Empirical Methods in Natural Language Processing and the 9th International Joint Conference on Natural Language Processing (EMNLP-IJCNLP)",
  year = "2019",
  pages = "3982--3992",
  publisher = "Association for Computational Linguistics",
  doi = "10.18653/v1/D19-1410",
  url = "https://aclanthology.org/D19-1410/"
}

@inproceedings{pei2023annotator,
  title={When do annotator demographics matter? measuring the influence of annotator demographics with the POPQUORN dataset},
  author={Pei, Jiaxin and Jurgens, David},
  booktitle={Proceedings of the 17th linguistic annotation workshop (LAW-XVII)},
  pages={252--265},
  year={2023}
}

@article{hu2022lora,
  title={Lora: Low-rank adaptation of large language models.},
  author={Hu, Edward J and Shen, Yelong and Wallis, Phillip and Allen-Zhu, Zeyuan and Li, Yuanzhi and Wang, Shean and Wang, Liang and Chen, Weizhu and others},
  journal={Iclr},
  volume={1},
  number={2},
  pages={3},
  year={2022}
}

@inproceedings{leonardelli2023semeval,
  title={SemEval-2023 task 11: Learning with disagreements (LeWiDi)},
  author={Leonardelli, Elisa and Abercrombie, Gavin and Almanea, Dina and Basile, Valerio and Fornaciari, Tommaso and Plank, Barbara and Rieser, Verena and Uma, Alexandra and Poesio, Massimo},
  booktitle={Proceedings of the 17th International Workshop on Semantic Evaluation (SemEval-2023)},
  pages={2304--2318},
  year={2023}
}

@article{davani2022dealing,
  title={Dealing with disagreements: Looking beyond the majority vote in subjective annotations},
  author={Davani, Aida Mostafazadeh and D{\'\i}az, Mark and Prabhakaran, Vinodkumar},
  journal={Transactions of the Association for Computational Linguistics},
  volume={10},
  pages={92--110},
  year={2022},
  publisher={MIT Press One Rogers Street, Cambridge, MA 02142-1209, USA journals-info~…}
}

@inproceedings{tahaeibeyond,
  title={Beyond Consensus: Use of Demographics for Datasets that Reflect Annotator Disagreement},
  author={Tahaei, Narjes and Bergler, Sabine},
  year = "2025",
  booktitle={First Workshop on Bridging NLP and Public Opinion Research}
}

@article{ross2017measuring,
  title={Measuring the reliability of hate speech annotations: The case of the european refugee crisis},
  author={Ross, Bj{\"o}rn and Rist, Michael and Carbonell, Guillermo and Cabrera, Benjamin and Kurowsky, Nils and Wojatzki, Michael},
  journal={arXiv preprint arXiv:1701.08118},
  year={2017}
}

@article{kennedy2020constructing,
  title={Constructing interval variables via faceted rasch measurement and multitask deep learning: a hate speech application},
  author={Kennedy, Chris J and Bacon, Geoff and Sahn, Alexander and von Vacano, Claudia},
  journal={arXiv preprint arXiv:2009.10277},
  year={2020}
}

@article{zhang2026modeling,
  title={Modeling Human Perspectives with Socio-Demographic Representations},
  author={Zhang, Leixin and Coltekin, Cagri},
  journal={arXiv preprint arXiv:2604.18069},
  year={2026}
}

@inproceedings{beck2024sensitivity,
  title={Sensitivity, performance, robustness: Deconstructing the effect of sociodemographic prompting},
  author={Beck, Tilman and Schuff, Hendrik and Lauscher, Anne and Gurevych, Iryna},
  booktitle={Proceedings of the 18th Conference of the European Chapter of the Association for Computational Linguistics (Volume 1: Long Papers)},
  pages={2589--2615},
  year={2024}
}

@article{lutz2025prompt,
  title={The prompt makes the person (a): A systematic evaluation of sociodemographic persona prompting for large language models},
  author={Lutz, Marlene and Sen, Indira and Ahnert, Georg and Rogers, Elisa and Strohmaier, Markus},
  journal={arXiv preprint arXiv:2507.16076},
  year={2025}
}

@article{touvron2023llama,
  title={Llama 2: Open foundation and fine-tuned chat models},
  author={Touvron, Hugo and Martin, Louis and Stone, Kevin and Albert, Peter and Almahairi, Amjad and Babaei, Yasmine and Bashlykov, Nikolay and Batra, Soumya and Bhargava, Prajjwal and Bhosale, Shruti and others},
  journal={arXiv preprint arXiv:2307.09288},
  year={2023}
}

@article{grattafiori2024llama,
  title={The llama 3 herd of models},
  author={Grattafiori, Aaron and Dubey, Abhimanyu and Jauhri, Abhinav and Pandey, Abhinav and Kadian, Abhishek and Al-Dahle, Ahmad and Letman, Aiesha and Mathur, Akhil and Schelten, Alan and Vaughan, Alex and others},
  journal={arXiv preprint arXiv:2407.21783},
  year={2024}
}

@article{Yang2024Qwen2TR,
  title={Qwen2 Technical Report},
  author={An Yang and Baosong Yang and Binyuan Hui and Bo Zheng and Bowen Yu and Chang Zhou and Chengpeng Li and Chengyuan Li and Dayiheng Liu and Fei Huang and Guanting Dong and Haoran Wei and Huan Lin and Jialong Tang and Jialin Wang and Jian Yang and Jianhong Tu and Jianwei Zhang and Jianxin Ma and Jin Xu and Jingren Zhou and Jinze Bai and Jinzheng He and Junyang Lin and Kai Dang and Keming Lu and Ke-Yang Chen and Kexin Yang and Mei Li and Min Xue and Na Ni and Pei Zhang and Peng Wang and Ru Peng and Rui Men and Ruize Gao and Runji Lin and Shijie Wang and Shuai Bai and Sinan Tan and Tianhang Zhu and Tianhao Li and Tianyu Liu and Wenbin Ge and Xiaodong Deng and Xiaohuan Zhou and Xingzhang Ren and Xinyu Zhang and Xipin Wei and Xuancheng Ren and Yang Fan and Yang Yao and Yichang Zhang and Yunyang Wan and Yunfei Chu and Zeyu Cui and Zhenru Zhang and Zhi-Wei Fan},
  journal={ArXiv},
  year={2024},
  volume={abs/2407.10671},
  url={https://api.semanticscholar.org/CorpusID:271212307}
}

@article{Yang2024Qwen25TR,
  title={Qwen2.5 Technical Report},
  author={Qwen An Yang and Baosong Yang and Beichen Zhang and Binyuan Hui and Bo Zheng and Bowen Yu and Chengyuan Li and Dayiheng Liu and Fei Huang and Guanting Dong and Haoran Wei and Huan Lin and Jian Yang and Jianhong Tu and Jianwei Zhang and Jianxin Yang and Jiaxin Yang and Jingren Zhou and Junyang Lin and Kai Dang and Keming Lu and Keqin Bao and Kexin Yang and Le Yu and Mei Li and Mingfeng Xue and Pei Zhang and Qin Zhu and Rui Men and Runji Lin and Tianhao Li and Tingyu Xia and Xingzhang Ren and Xuancheng Ren and Yang Fan and Yang Su and Yi-Chao Zhang and Yunyang Wan and Yuqi Liu and Zeyu Cui and Zhenru Zhang and Zihan Qiu and Shanghaoran Quan and Zekun Wang},
  journal={ArXiv},
  year={2024},
  volume={abs/2412.15115},
  url={https://api.semanticscholar.org/CorpusID:274859421}
}

@article{Jiang2023Mistral7,
  title={Mistral 7B},
  author={Albert Qiaochu Jiang and Alexandre Sablayrolles and Arthur Mensch and Chris Bamford and Devendra Singh Chaplot and Diego de Las Casas and Florian Bressand and Gianna Lengyel and Guillaume Lample and Lucile Saulnier and L{\'e}lio Renard Lavaud and Marie-Anne Lachaux and Pierre Stock and Teven Le Scao and Thibaut Lavril and Thomas Wang and Timoth{\'e}e Lacroix and William El Sayed},
  journal={ArXiv},
  year={2023},
  volume={abs/2310.06825},
  url={https://api.semanticscholar.org/CorpusID:263830494}
}

\appendix

\begin{table*}[t]
\centering
\setlength{\tabcolsep}{4.5pt}
\begin{tabular}{lllcccccc}
\toprule
Split & Framework & Input setting
& Acc. & Prec. & Rec. & F1 & AUC & $\Delta$AUC \\
\midrule

\multirow{6}{*}{Split A}
& \multirow{3}{*}{Framework A}
& Text only
& 61.28 & 59.99 & 59.40 & 59.39 & 64.73 & -- \\
& & + Shuffled demographics
& 60.99 & 59.80 & 59.48 & 59.51 & 64.54 & \textbf{\textcolor{negRed}{-0.19}} \\
& & + Real demographics
& \textcolor{posGreen}{61.91} & \textcolor{posGreen}{60.74} & \textcolor{posGreen}{60.22} & \textcolor{posGreen}{60.21} & \textcolor{posGreen}{66.01} & \textbf{\gain{1.28}} \\

\cmidrule(lr){2-9}

& \multirow{3}{*}{Framework B}
& Text only
& 72.19 & \textcolor{posGreen}{69.81} & 69.41 & 69.43 & 78.03 & -- \\
& & + Shuffled demographics
& 71.75 & 69.56 & \textcolor{posGreen}{69.90} & 69.55 & 77.93 & \textbf{\textcolor{negRed}{-0.10}} \\
& & + Real demographics
& \textcolor{posGreen}{72.21} & 69.79 & 69.57 & \textcolor{posGreen}{69.58} & \textcolor{posGreen}{78.21} & \textbf{\gain{0.18}} \\

\midrule

\multirow{6}{*}{Split B}
& \multirow{3}{*}{Framework A}
& Text only
& \textcolor{posGreen}{81.76} & 82.84 & \textcolor{posGreen}{81.68} & \textcolor{posGreen}{81.57} & \textcolor{posGreen}{92.26} & -- \\
& & + Shuffled demographics
& 81.06 & 82.46 & 80.97 & 80.80 & 91.91 & \textbf{\textcolor{negRed}{-0.35}} \\
& & + Real demographics
& 81.40 & \textcolor{posGreen}{82.89} & 81.30 & 81.13 & 92.25 & \textbf{\textcolor{negRed}{-0.01}} \\

\cmidrule(lr){2-9}

& \multirow{3}{*}{Framework B}
& Text only
& \textcolor{posGreen}{82.22} & \textcolor{posGreen}{81.22} & 80.23 & 80.57 & \textcolor{posGreen}{90.41} & -- \\
& & + Shuffled demographics
& 81.79 & 80.53 & 80.30 & 80.32 & 90.09 & \textbf{\textcolor{negRed}{-0.32}} \\
& & + Real demographics
& 82.11 & 80.92 & \textcolor{posGreen}{80.91} & {80.76} & 90.40 & \textbf{\textcolor{negRed}{-0.01}} \\

\bottomrule
\end{tabular}
\caption{
Effect of demographic information across the full split--framework comparison.
For each split--framework setting, $\Delta$AUC is computed relative to the text-only baseline.
Shuffled demographics serve as a control for whether gains come from real demographic signal rather than additional feature dimensions.
Best scores within each three-row input-setting block are highlighted in bold brick red.
}
\label{tab:full_demo_split_framework}
\end{table*}


\begin{figure*}[t]
\centering
\begin{subfigure}[t]{0.24\textwidth}
    \centering
    \includegraphics[width=\linewidth]{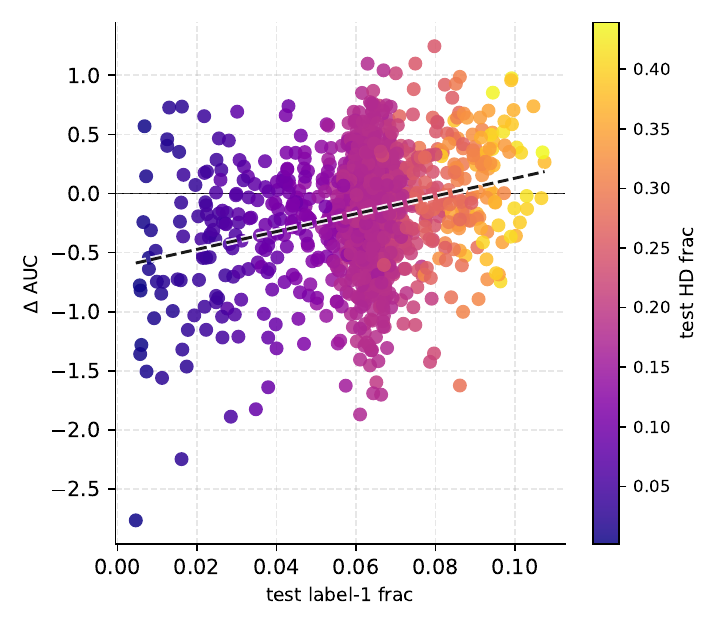}
    \caption{Ambiguity level of test}
    \label{fig:delta_auc_test_label_frac}
\end{subfigure}
\hfill
\begin{subfigure}[t]{0.24\textwidth}
    \centering
    \includegraphics[width=\linewidth]{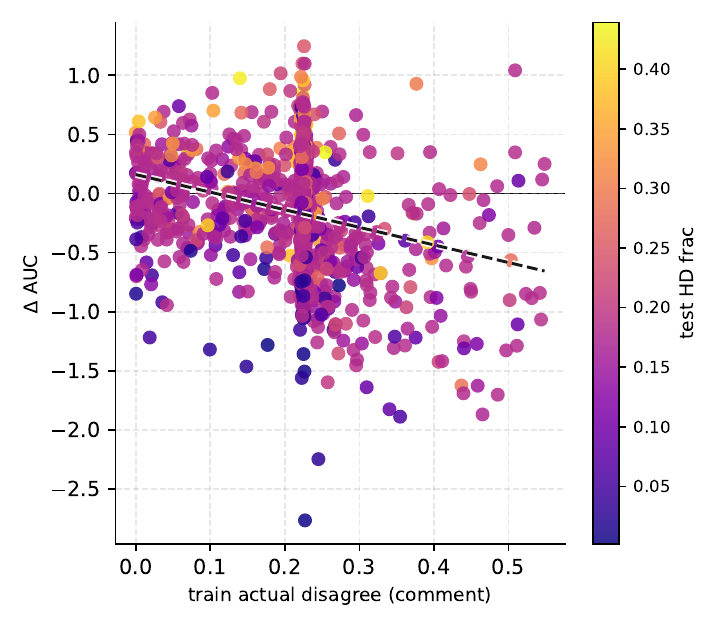}
    \caption{Ambiguity level of train}
    \label{fig:delta_auc_train_disagree}
\end{subfigure}
\hfill
\begin{subfigure}[t]{0.24\textwidth}
    \centering
    \includegraphics[width=\linewidth]{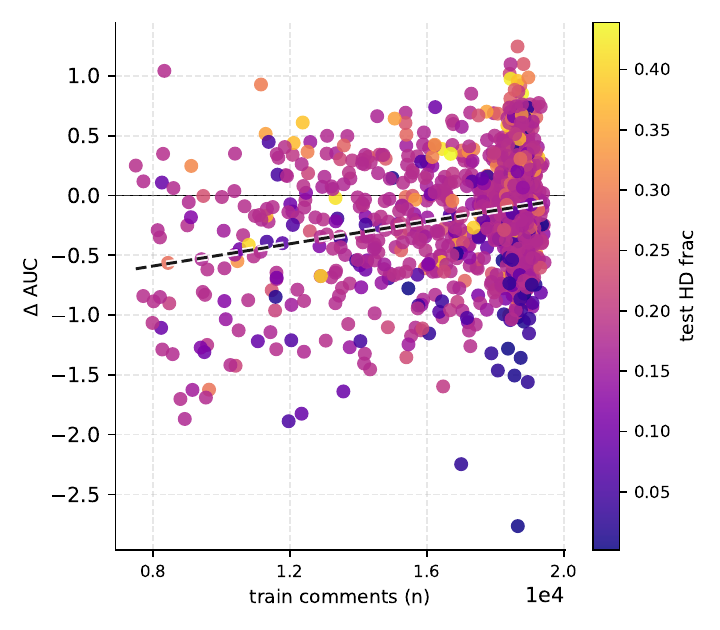}
    \caption{Scale of train}
    \label{fig:delta_auc_train_size}
\end{subfigure}
\hfill
\begin{subfigure}[t]{0.24\textwidth}
    \centering
    \includegraphics[width=\linewidth]{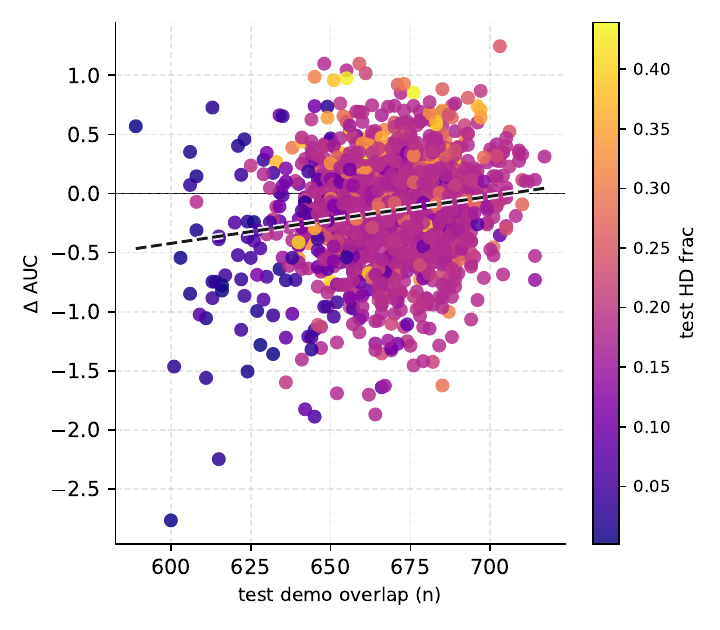}
    \caption{Demographic coverage}
    \label{fig:delta_auc_demo_overlap}
\end{subfigure}

\caption{
Factors associated with the AUC gain from adding demographic features.
We examine test uncertain-label fraction, training disagreement, training size, and test demographic overlap.
Together, these results characterize when demographic features provide larger or smaller gains.
}
\label{fig:demo_gain_factors}
\end{figure*}


\section{Case Study: When Do Demographics Help?}
\label{app:intro_case}
We illustrate two representative data splits with different disagreement profiles. 
Split A has lower training disagreement but much higher test disagreement, with comment-level disagreement means of 0.2059 in train and 0.7461 in test. 
Split B has more similar train and test disagreement, with means of 0.2642 and 0.2379, respectively.

We evaluate two demographic-aware modeling frameworks. Framework A uses a frozen Sentence-BERT text encoder, concatenates its text representation with learnable demographic embeddings, and trains an MLP classifier. Framework B finetunes BERTweet with learnable demographic special tokens prepended to the text.

We also include a shuffled-demographic baseline as a negative control for the effect of adding an extra demographic input channel. The shuffled features preserve the demographic feature distribution but break the true annotator-label correspondence. 
Therefore, any additional gain from real demographics over shuffled demographics indicates that the model is using meaningful demographic signal rather than simply benefiting from extra input capacity or global bias. Details results for Table~\ref{tab:demo_split_framework_summary} can be found in Table~\ref{tab:full_demo_split_framework}.


\section{Additional Split-Descriptor Analysis}
\label{app:split_descriptor_plots}
Figure~\ref{fig:demo_gain_factors} visualizes the individual split descriptors summarized in Table~\ref{tab:demo_gain_correlations}. 
These plots provide additional descriptive evidence that demographic gains tend to be larger when test ambiguity is higher, training disagreement is lower, training data are larger, and train--test demographic overlap is greater.


\section{Multivariate Analysis of Split-Level Factors}
\label{app:multivariate-split-regression}
To further validate the split regimes discussed in Section~\ref{sec:regimes_split}, we conduct a multivariate regression analysis on the MHS three-class splits. 
The dependent variable is the performance gain from adding demographic information, measured as $\Delta \mathrm{AUC}=\mathrm{AUC}_{\text{text+demo}}-\mathrm{AUC}_{\text{text}}$. 
We include four split-level predictors: mean training disagreement, mean test disagreement, training set size, and demographic-combination coverage, measured as the number of test demographic combinations observed in training. 
All predictors and the dependent variable are $z$-standardized before regression, so the coefficients can be interpreted as standardized effect sizes.

Table~\ref{tab:appendix-multivariate-regression} supports our findings. 
After controlling for the other split characteristics, training disagreement remains the strongest negative predictor of demographic gain, whereas test disagreement remains a strong positive predictor. 
This indicates that demographic information is most useful when the evaluation set contains ambiguous examples, but high disagreement in the training set can make learning noisier. 
Training size also has a positive association with demographic gain, suggesting that more data helps stabilize the use of demographic information and gain more information over the text-only model. 
Demographic overlap is positive and significant, but its standardized effect is smaller than the disagreement variables, indicating that demographic coverage matters.
Overall, the multivariate analysis reinforces our conclusion that the usefulness of demographic information is primarily influenced by proposed data regimes.


\begin{table*}[t]
\centering
\begin{tabular}{lccc}
\toprule
Predictor & Std. Coef. & $p$-value & VIF \\
\midrule
Train disagreement mean & -0.292 & $<.001$ & 1.02 \\
Test disagreement mean & 0.223 & $<.001$ & 1.13 \\
Train records & 0.148 & $<.001$ & 1.03 \\
Test demographic overlap count & 0.086 & $<.001$ & 1.15 \\
\bottomrule
\end{tabular}
\caption{Multivariate regression predicting $\Delta$AUC on MHS three-class splits. All variables are $z$-standardized. The model explains $R^2=0.188$ of the variance over 1,400 sampled splits. Low VIF values indicate little multicollinearity among predictors.}
\label{tab:appendix-multivariate-regression}
\end{table*}

\begin{table*}[t]
\centering
\small
\begin{tabular}{llrrrrrrrr}
\toprule
Dataset & Split & \#Ann. & \#Comments & \#Annot. & Ann./Com. & $y=0/y=1$ & $y{=}1$ (\%) & $D_{\text{orig}}$ & $D_{\text{bin}}$ \\
\midrule
MHS & Train & 72,130 & 27,319 & 5,538 & 2.64 & 44,756 / 27,374 & 38.0 & 0.265 & 0.226 \\
MHS & Valid & 7,558 & 6,201 & 2,109 & 1.22 & 5,386 / 2,172 & 28.7 & 0.169 & 0.084 \\
MHS & Test  & 7,596 & 6,242 & 2,127 & 1.22 & 5,400 / 2,196 & 28.9 & 0.169 & 0.081 \\
\midrule
POPQUORN & Train & 9,098 & 1,050 & 262 & 8.66 & 6,790 / 2,308 & 25.4 & 0.522 & 0.573 \\
POPQUORN & Valid & 1,969 & 225 & 262 & 8.75 & 1,428 / 541 & 27.5 & 0.533 & 0.576 \\
POPQUORN & Test  & 1,969 & 225 & 261 & 8.75 & 1,488 / 481 & 24.4 & 0.523 & 0.580 \\
\bottomrule
\end{tabular}
\caption{
Basic statistics of the binary MHS and POPQUORN datasets used in our experiments.
\#Ann. denotes annotation-level instances, \#Comments denotes unique comments, and \#Annot. denotes unique annotators.
Ann./Com. is the average number of annotations per comment, and $y{=}1$ (\%) reports the positive label proportion.
For MHS, we construct the binary task by removing the original middle class and remapping the remaining labels to
$0$ and $1$. For POPQUORN, we map the original 1--5 offensiveness ratings into binary labels.
$D_{\text{orig}}$ denotes normalized comment-level disagreement computed from the original label space
(MHS: three classes; POPQUORN: five offensiveness ratings), while $D_{\text{bin}}$ denotes disagreement recomputed
after binarization.
}
\label{tab:dataset_statistics}
\end{table*}

\section{Experiment Details}
\label{app:exp_details}
We report the hyperparameters of the residual model in Table~\ref{tab:best_gate_hyperparams} and provide module-level analysis in Section~\ref{sec:analysis_residual_model}. 
The residual formulation provides the main benefit, while the gate and soft-label components offer flexibility for different data regimes and evaluation goals. 
We report accuracy, macro precision, macro recall, macro F1, and ROC-AUC.
All results represent the average of five runs.

\paragraph{Use of artifacts.}
We use publicly available datasets and pretrained models only for scientific research.
We cite the original creators of all datasets and model checkpoints used in our experiments, and follow the access conditions and terms provided by their original releases.

\subsection{Dataset Details}
\label{app:dataset}

Table~\ref{tab:dataset_statistics} reports the basic statistics of the two binary datasets used in our experiments. 
We use MHS as the main dataset for analyzing demographic-favorable regimes.
We do not use POPQUORN for this regime discovery analysis because POPQUORN was intentionally sampled by its original authors to focus on high-disagreement cases, creating a setting in which annotators are more likely to diverge in their judgments. 
It is therefore less suitable for studying how demographic gains vary across low- and high-disagreement regimes. 
In addition, POPQUORN contains only 1,500 unique comments, which makes text-only generalization more difficult and can increase the relative value of demographic signals.

Binarization affects disagreement differently in the two datasets. In POPQUORN, binary disagreement increases because many fine-grained rating differences lie near the offensiveness threshold. 
Across all splits, 74.8\% of comments contain ratings on both sides of the binary boundary (1--2 vs. 3--5), and 59.7\% of comments have higher disagreement after binarization. 
Among boundary-crossing comments, 64.3\% contain both ratings 2 and 3, suggesting that the increase is mainly driven by ambiguity around the non-offensive/offensive decision boundary.
By contrast, binary disagreement decreases in MHS because the intermediate class
is removed rather than merged into either binary class. 
Many originally ambiguous comments involve this removed class: in the test split, 45.7\% of originally disagreeing comments contain class 1, and 43.3\% involve only 0/1 or 1/2 disagreement. After removing class 1, such comments often become single-sided
and no longer contribute to binary disagreement.

\subsection{Prompt Templates}
\label{app:prompt_template}
We use two prompt formats for LLM-based baselines. For zero-shot prompting and
soft-prompt tuning, we use instruction-style prompts that define the task and
constrain the output to a single numeric label. For LoRA finetuning, we use a
minimal input template and train the model with a classification head. In all
templates, \(\{x\}\) denotes the input text and \(\{d\}\) denotes the serialized
annotator demographic attributes.

\paragraph{Zero-shot and soft-prompt templates.}
For MHS, we use the following text-only prompt:
\begin{quote}
\small
Hate speech is defined as language that attacks or demeans a person or group
based on attributes like race, religion, gender, etc. Use this binary label set:
0 = not hateful, 1 = hateful. Based on the given definition, determine how people
are likely to label the following text. Text: \(\{x\}\). Answer with only one
number: 0 or 1. Answer:
\end{quote}

The demographic-aware version adds annotator background information:
\begin{quote}
\small
Hate speech is defined as language that attacks or demeans a person or group
based on attributes like race, religion, gender, etc. Personal background:
\(\{d\}\). Use this binary label set: 0 = not hateful, 1 = hateful. Based on the
given definition and the provided personal background information, determine how
the annotator is likely to label the following text. Text: \(\{x\}\). Answer
with only one number: 0 or 1. Answer:
\end{quote}
We extract the logits corresponding to ``0'' and ``1'' to serve as the prediction results.

For POPQUORN, we use an offensiveness-specific prompt:
\begin{quote}
\small
Offensiveness is the degree to which a Reddit comment is insulting, rude,
hurtful, or otherwise offensive to a reader. Use this binary label set: 0 = not
offensive, 1 = offensive. Determine how people are likely to label the following
Reddit comment. Text: \(\{x\}\). Answer with only one number: 0 or 1. Answer:
\end{quote}

The demographic-aware POPQUORN prompt is:
\begin{quote}
\small
Offensiveness is the degree to which a Reddit comment is insulting, rude,
hurtful, or otherwise offensive to a reader. Personal background: \(\{d\}\). Use
this binary label set: 0 = not offensive, 1 = offensive. Based on the provided
personal background information, determine how the annotator is likely to label
the following Reddit comment. Text: \(\{x\}\). Answer with only one number: 0 or
1. Answer:
\end{quote}

\paragraph{LoRA finetuning templates.}
For LoRA-based finetuning, the text-only input is simply \(\{x\}\).
The demographic-aware input is:
\begin{quote}
\small
Annotator: \(\{d\}\) \\
Text: \(\{x\}\)
\end{quote}

Here, \(\{d\}\) is a comma-separated list of available demographic attributes.
Unlike the zero-shot setting, the LoRA model is not asked to generate the label
string. Instead, the input is passed to a sequence-classification head trained
with the binary supervision signal.

\section{Analysis of Gated Residual Model}
\label{sec:analysis_residual_model}
Table~\ref{tab:mhs_gate_ablation} reports ablation results on MHS with BERTweet. 
(1) Together with Table~\ref{tab:main_results_compact}, the ablation suggests that the main gain comes from modeling demographics as an explicit residual correction to a text-only classifier. 
Instead of asking the model to learn text and demographic signals in a single fused representation, the residual design gives demographic information a direct pathway to adjust the text-only prediction. 
This reduces the risk that demographic signals are overwhelmed by the much stronger textual signal, while keeping their role interpretable as a correction rather than a replacement;
(2) within the residual framework, the gate and soft-label loss mainly affect how the demographic correction is used. 
The full model achieves the best recall and F1, while variants without the gate or without soft-label supervision obtain slightly higher accuracy, precision, or AUC. 
Soft-label supervision encourages the text-only model to produce smoother uncertainty estimates that reflect aggregate annotator disagreement, and the gate emphasizes examples where the text-only model is less confident. 
These components help model become less overly conservative in the MHS binary setting, where labels are imbalanced and disagreement is compressed, to recover more positive cases, leading to higher recall and F1;
(3) these results show that the best configuration depends on the data regime and target metric. 
Removing the gate slightly improves AUC in this setting, but reduces recall and F1. 
The full model provides a better balance for threshold-based classification, while maintaining comparable AUC. 
Overall, the ablation supports our view that demographic conditioning should be treated as a flexible residual adjustment whose use can be tuned according to the dataset and evaluation goal.

\begin{table*}[t]
\centering
\begin{tabular}{lccccc}
\toprule
Model & Acc. & Prec. & Rec. & F1 & AUC \\
\midrule
Full model
& \textit{81.58}
& 77.71
& \textbf{77.00}
& \textbf{77.31}
& 87.39 \\
w/o gate ($\alpha=1.0, \rho=0.0$)
& \textbf{81.64}
& \textbf{78.00}
& 76.12
& 76.92
& \textbf{87.50} \\
w/o gate-weighted loss ($\rho=0.0$)
& 81.54
& \textit{77.88}
& 76.25
& 76.87
& \textit{87.43} \\
w/o soft-label loss ($\lambda_s=0.0$)
& \textbf{81.64}
& 77.87
& \textit{76.70}
& \textit{77.19}
& 87.35 \\
\bottomrule
\end{tabular}
\caption{Ablation results on MHS with BERTweet. Results are averaged over five runs.}
\label{tab:mhs_gate_ablation}
\end{table*}

\begin{table*}[t]
\centering
\begin{tabular}{llccccccc}
\hline
Dataset & Backbone 
& $\rho$ & $\tau$ & $T$
& $\lambda_s$
& $E_t$ & $E_r$
& Dropout \\
\hline
MHS & BERTweet
& 0.50 & 0.55 & 0.08
& 0.50
& 10 & 20
& 0.20 \\

MHS & ToxDect-RoBERTa
& 0.40 & 0.50 & 0.12
& 0.50
& 10 & 20
& 0.25 \\

POPQUORN & BERTweet
& 0.25 & 0.45 & 0.15
& 0.20
& 15 & 30
& 0.25 \\

POPQUORN & ToxDect-RoBERTa
& 0.25 & 0.55 & 0.15
& 0.50
& 10 & 20
& 0.20 \\
\hline
\end{tabular}
\caption{
Best hyperparameters for the gated demographic residual model.
$\rho$ is the residual cross-entropy gate weight, $\tau$ is the uncertainty gate threshold,
$T$ is the gate temperature, $\lambda_s$ is the soft-label loss weight,
$E_t$ and $E_r$ are PLMs teacher and residual training epochs, $\eta_r$ is the residual learning rate.
Shared settings: PLMs teacher learning rate $1{\times}10^{-5}$, PLMs teacher head learning rate $1{\times}10^{-3}$, categorical embedding dimension $64$,
binary embedding dimension $16$, residual hidden size $256$, and validation selection by AUC.
}
\label{tab:best_gate_hyperparams}
\end{table*}

\end{document}